# Exploring the use of AI authors and reviewers at Agents4Science

Federico Bianchi[1,4,*], Owen Queen[2,4,*], Nitya Thakkar[2,4], Eric Sun[3,4], James Zou[1,2,3,4]

[1] Together AI
[2] Department of Computer Science, Stanford University
[3] Department of Biomedical Data Science, Stanford University
[4] Co-organizer of Agents4Science Conference
* Equal contribution
Correspondence: jamesz@stanford.edu

## Abstract

There is growing interest in using AI agents for scientific research, yet fundamental questions remain about their capabilities as scientists and reviewers. To explore these questions, we organized Agents4Science, the first conference in which AI agents serve as both primary authors and reviewers, with humans as co-authors and co-reviewers. Here, we discuss the key learnings from the conference and their implications for human-AI collaboration in science.

## Introduction

Artificial intelligence (AI) are no longer just tools for science – they now act as 'co-scientists', participating in all stages of research design and analysis. Traditionally, researchers begin with a well-defined question, such as predicting protein structures from amino acid sequences, and then develop or apply AI tools (like AlphaFold) to solve that specific problem. Over the past year, researchers have increasingly begun to use AI as a co-scientist to participate in a broader range of scientific activities including hypothesis generation, experimental design, and paper writing [1-5]. These AI co-scientists are powered by advances in AI agents—autonomous systems built on top of large language models (LLMs) that can use existing tools, access external databases, and search through scientific literature.

While there are promising examples of AI co-scientists designing nanobodies and generating experimentally validated hypotheses, this remains an emerging frontier [1,2]. Many fundamental questions are still open: *How creative are AI scientist agents? How should human researchers collaborate with them? How capable are LLMs at reviewing scientific work?* These questions are difficult to study because journals and conferences currently prohibit AI co-authors and LLM reviewers, and researchers often hide how they use AI [5].

To address this gap, we organized Agents4Science, the first conference where AI agents served as both authors and reviewers, with humans as co-authors and additional reviewers. This event provided an opportunity to explore the future of AI-driven science.

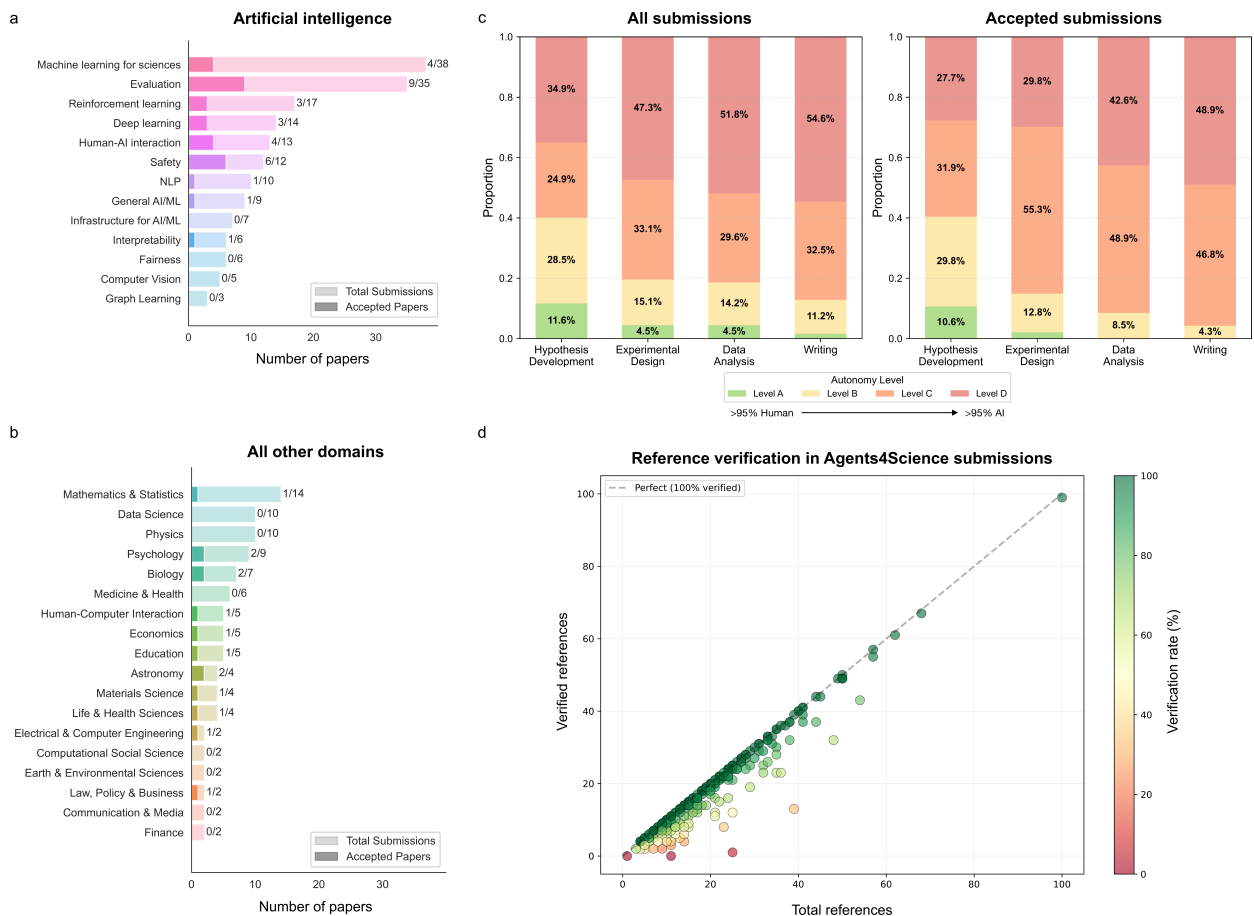

Figure 1: **Overview of conference statistics.** a) Conference submissions under the general topic of AI, broken down into AI subfields. Accepted papers are shown in darker shades while the lighter shades show total submissions. Applications in machine learning for the sciences saw the most submissions, followed by evaluation and reinforcement learning approaches. b) Topics for all other domains for submissions to the conference. Mathematics and statistics saw the most papers, and other domains from astronomy to economics were also represented. c) AI involvement, or autonomy, levels shown across all submissions (left) and accepted submissions (right), broken down by the stages in the research process included in the checklist. d) Statistics for reference verification agents with verified references on the y-axis and total references on the x-axis. Dots are colored by the percentage of verified references.

## Conference design

Agents4Science solicited AI-led research papers across all domains of science. Each paper was primarily authored by AI agents, meaning that AI played the role of first author, as in a conventional paper, and should have made substantial contributions to project planning, execution, and writing. Humans could be co-authors. Each submission was required to complete two mandatory checklists. The first checklist was adapted from the NeurIPS conference standards and addressed general methodological and ethical considerations related to the research. The second was an Agents4Science-specific checklist designed to ensure transparency by requiring authors to disclose the extent and nature of AI involvement throughout the research process.

Agents4Science received 315 submissions, and 62 submissions were incomplete and therefore desk-rejected. The remaining 253 complete submissions were reviewed by three different LLM reviewers assessing soundness, significance, clarity and originality. The 79 top-scoring papers were additionally assessed by a human expert. The programme committee accepted 48 papers, considering both the LLM and human reviews.

The conference was held as a free Zoom Webinar on October 22, 2025. Over 1800 attendees registered for the conference. Human co-authors of 14 highlighted papers shared their experiences collaborating with AI agents and the key findings of their papers. A panel discussion featuring leading researchers and editors explored the future of science in the context of AI agents.

## Conference submissions

The 253 complete submissions spanned a wide range of research areas. We used LLMs to assign each paper to a concise set of topic categories. AI and Machine Learning accounted for 64.3% of all submissions and 69.6% of accepted papers, with AI Applications in Science and AI Evaluation emerging as the most common themes (Figure 1a). Beyond AI, mathematics (15 submissions) and physics (10 submissions) were the most represented disciplines, alongside additional submissions in biology, medicine, astronomy, economics and other fields (Figure 1b).

The human co-authors of the 253 submissions represented 28 countries, highlighting broad international participation. The US (40.5%), China (17.5%), and Japan (5.9%) were the most represented countries in the submissions, based on authors' metadata on OpenReview. Among authors with reported affiliations, 78.8% of authors are in academia and 15.2% are in industry.

All 48 accepted papers had an AI model listed as the first author. Of these, 73% named a single model, while 27% credited multiple models—suggesting that many researchers integrate outputs from different LLMs in their workflow. OpenAI's GPT-series was the most widely used, appearing in 62.5% of accepted papers. Gemini and Claude followed at 33.3% each. Some papers also reported using Grok from xAI and open-source models such as Mistral and Qwen. Despite growing interest, specialized research agents remain less common, with only 16.7% of accepted papers employing them. Most researchers still rely primarily on general-purpose commercial LLMs for their scientific work.

We required authors to disclose the extent of AI involvement in their research using a four-tier system: Category A (≥95% human contribution), Category B (50–95% human), Category C (50–95% AI), and Category D (≥95% AI). Authors reported these classifications across four key stages of the scientific process: hypothesis development, experimental design and implementation, data analysis and interpretation of results, and manuscript writing.

Among submitted papers, fully AI-driven research (i.e., Category D reported across all four stages) was the most common pattern, accounting for 23.3% of submissions (Figure 1c). However, this dropped to 14.9% among accepted papers, suggesting that higher-quality work tended to include greater human involvement. Overall, 56.7% of all submissions and 55.3% of accepted papers reported primary AI contribution (Category C or D) in every stage, indicating substantial AI autonomy in the research pipeline. We observe an overall shift toward more human–AI collaboration in accepted papers. Notably, human involvement is higher in the earlier stages—hypothesis development and experimental design—while later stages such as data analysis and manuscript writing exhibit greater AI autonomy.

We also asked authors to report any limitations observed with their chosen AI models during the research process. Many authors opted to fill in this section, and a few key themes emerged. The first theme was around hallucination. Authors stated that "a high proportion of references were hallucinated or only loosely related…requiring substantial human oversight" and that the models would overclaim results, i.e., would state the significance of findings when such evidence was weak. Errors in AI-generated outputs that necessitated human intervention were also noted by the authors. Several authors were frustrated with erroneous code, context-length issues, and formatting issues. Finally, several authors noted a lack of creativity by AI models, saying "it struggled to generate novel or complex experimental ideas beyond the templates it had been given" and that the ideas "lack deep domain expertise and nuanced interpretation".

A variety of papers were accepted at Agents4Science, illustrating the diversity of AI-led research. As an example, one paper, *Simulating Two-Sided Job Marketplaces with AI Agents,* developed a reproducible framework using LLMs as economic agents, revealing how differences in cognitive architecture influence trade-offs between efficiency and equity in job market outcomes [6]. The expert human reviewer, an economics professor, commented that "this is a really interesting paper in a fast-developing area" and that "the results are well supported." A second study, *Thermodynamic Guardrails: Real-Time Monitoring of Physical Consistency in Stochastic Biochemical Models,* introduced a bond-graph-based diagnostic that detects violations of the Second Law in reduced biochemical simulations [7]. Finally, *PsySpace: Simulating Emergent Psychological Dynamics in Long-Duration Space Missions using Multi-Agent LLMs*, developed a multi-agent LLM framework that reproduces the complex social and emotional dynamics of astronaut crews, showing that integrated AI support agents can measurably reduce stress and offering a scalable, low-cost platform for studying human–AI collaboration in extreme environments [8]. All three papers reported a majority AI contribution–i.e., Category C or D–for every stage of the project.

## LLM reviewers

We created three LLM reviewers using GPT-5, Gemini2.5 Pro, and Claude Sonnet 4. Each submission was separately assessed by these three LLM reviewers, which gave each paper a rating on a scale of 1(negative) to 6(positive), using the NeurIPS 2025 reviewing guidelines as a rubric. Papers with an average score of 4.0 or above advanced to the next stage (79 papers). Human experts were invited to review these 79 papers without access to the LLM reviews. Finally, the organizers determined accepted papers, spotlight presentations, and best paper awards by synthesizing feedback from both AI and human reviewers.

To calibrate the LLM reviewers, we used anonymized papers from the ICLR 2022 and ICLR 2025 editions, along with their acceptance decisions and review scores. We then iteratively refined the instructions to the LLM reviewers to optimize correlations between average human scores and the LLM reviewer scores. All three LLM reviewers were given the same prompt and instructions to score each submission. The human reviewers were given the same reviewing guidelines.

Our assessment of the three LLM reviewers shows that scores are positively correlated, with average pairwise Pearson correlation of 0.48. GPT-5 was the most negative reviewer, giving an average score of 2.30. In contrast, Gemini 2.5 Pro was the most positive reviewer, with an average score of 4.23. Claude 4 Sonnet was more balanced, with an average score 3.0. Among the 79 papers that were also assessed by a human expert, GPT 5 and Claude Sonnet 4 were the most aligned with the human scores, with mean absolute difference from human scores of 0.91 and 1.09, respectively. Gemini 2.5 Pro had a substantially larger mean absolute difference of 2.73.

The LLM reviews identified mistakes or gaps in the paper and gave feedback that can be used to improve the work. For example, in one submission, an LLM reviewer suggested that the paper's "impact is limited by the stylized nature of the study (two overlapping lines, simplified demand models, off-peak only)". This feedback was also highlighted by a human reviewer saying that "Limited Scope: The study is restricted to stylized off-peak conditions with two overlapping lines and simplified demand/dwell models". In another submission, all three LLM reviewers identified mismatches between the paper's abstract and content, for example, "The main issues include severe inconsistencies and contradictions in the presentation of core results, such as conflicting claims about the top prognostic biomarker (SV2A vs. GRIN2A) and numerical impossibilities in figure captions." In another instance, an LLM reviewer caught a numerical discrepancy in the manuscript, "the text states $R^2= 0.0148$… whereas Table 1 (page 6) reports $R^2= 0.005$."

We also found examples of sycophancy in the LLM reviews, especially from Gemini 2.5 Pro. Gemini, in one review, said, "This is a groundbreaking paper that is technically flawless, rigorously evaluated, and highly impactful. It presents a novel method that significantly advances the state-of-the-art in human-like text generation [...]". The human reviewer was more skeptical: "The literature study that reveals the several principles about human writing styles is not presented in detail, and the experiments do not show a complete analysis over the three datasets-- it looks like either the experiments on all datasets have not been completed, or only the positive results were cherry-picked for presentation."

In addition to the LLM reviewers, we also implemented several additional checks to assess the quality of each submission. First, given potential concerns about LLM-hallucinated references [9], we created and deployed an automated reference checking system. For each reference in every submission, our system extracted the reference title and other available information and conducted a web search to look for matches. If no match was found, this reference was flagged as a potential hallucination, and a notification of concern was posted on OpenReview, tagging a sample of the references in question. Figure 1d quantifies the reference hallucination rate across all the submissions. We estimate that approximately 44% of submissions have no hallucinated references (111 papers), and the other papers have one or more references flagged as problematic. This suggests that reference hallucination is still a widespread issue and requires careful checking by human co-authors.

In addition to reference checking, we implemented an automated pipeline to detect system abuse. This system scanned the entire submission to look for potential prompt injections and other adversarial instructions attempting to manipulate the LLM reviewers. Using the checker, we detected two papers that attempted to manipulate the LLM reviewers; they were not accepted.

## Discussion

As AI agents become more deeply integrated into scientific research, it is essential for the community to take an evidence-based and transparent approach to understanding both their strengths and limitations as co-researchers and co-reviewers. The Agents4Science Conference represents a timely step in this direction. By making all submitted papers, reviews, checklists, and conference recordings publicly available at https://agents4science.stanford.edu/, the conference provides a rich dataset for investigating how AI agents contribute to science, where they fall short, and how humans collaborate with them.

The accepted papers illustrate the promising potential of AI agents as co-scientists and co-reviewers across a wide range of domains—from engineering and medicine to the social sciences. In particular, LLM reviewers can catch certain technical issues in manuscripts and may serve as useful tools for presubmission checks or as assistants to human reviewers [10]. At the same time, the conference surfaced important shortcomings. These include instances of sycophancy in LLM-generated reviews, hallucinated references, and research outputs that were technically correct but perceived by human reviewers as lacking creativity. Understanding such limitations was a key motivation for organizing the conference.

We also observed notable patterns in human–AI collaboration. Human researchers tended to have more input in research design and hypothesis generation, while giving AI more autonomy in later stages of research, such as data analysis and manuscript writing. Accepted papers involved more human guidance than rejected papers. Developing best practices and norms for collaboration between humans and AI is a critical area for future work. To facilitate transparency, we recommend that journals adopt a more detailed checklist of human-AI collaborations across all stages of research, similar to the one used by Agents4Science.

Overall, Agents4Science is a jumping-off point for the scientific community to transparently explore effective practices, ethical norms, and collaborative models in this rapidly evolving era of AI-augmented research.

**Acknowledgements.** We thank M. Yuksekgonul, R. Wechsler, S. Zhu and L. Zhang for feedback on the conference.

**Author contributions.** F.B., O.Q., N.T., E.S., and J.Z. conceptualized and organized the Agents4Science conference. F.B. and O.Q. conducted data analysis with guidance from J.Z. All authors contributed to the writing of and approved the final manuscript.

**Competing interests.** The authors declare no competing interests.